\documentclass[11pt,a4paper]{article}
\usepackage[hyperref]{naaclhlt2019}
\usepackage{times}
\usepackage{latexsym}

\usepackage{url}

\aclfinalcopy 
\def\aclpaperid{1331} 

\usepackage{latexsym}
\usepackage{pslatex}
\usepackage{latexsym}
\usepackage{xspace}
\usepackage{multirow}
\usepackage{amsfonts}
\usepackage{epstopdf}
\usepackage{amsmath}
\usepackage{caption}
\usepackage{tabularx}
\usepackage{subcaption}
\usepackage{pbox}
\usepackage{color}
\usepackage{xcolor}
\usepackage{microtype}
\usepackage{enumitem}  
\usepackage{booktabs}

\usepackage[colorinlistoftodos,prependcaption,textsize=scriptsize]{todonotes}

\newcommand{\dataset}[1]{#1\xspace}
\newcommand{\ourdataset}{\dataset{MultiSense}}
\newcommand{\multiflickr}{\dataset{Multi30k}}

\author{}

\title{Cross-lingual Visual Verb Sense Disambiguation}

\author{Spandana Gella$^{*}$ \and Desmond Elliott$^{\dagger}$ \and Frank Keller$^{*}$\\
  $^{*}$School of Informatics, University of Edinburgh\\
  $^{\dagger}$Department of Computer Science, University of Copenhagen\\
  \texttt{\{spandana.gella,frank.keller\}@ed.ac.uk}, \texttt{de@di.ku.dk}
}

\begin{document}

\maketitle

\begin{abstract}
  Recent work has shown that visual context improves cross-lingual sense disambiguation for nouns. We extend this line of work to the more challenging task of cross-lingual \textit{verb sense} disambiguation, introducing the \ourdataset dataset of 9,504 images annotated with English, German, and Spanish verbs. Each image in \ourdataset is annotated with an English verb and its translation in German or Spanish. We show that cross-lingual verb sense disambiguation models benefit from visual context, compared to unimodal baselines. We also show that the verb sense predicted by our best disambiguation model can improve the results of a text-only machine translation system when used for a multimodal translation task.
\end{abstract}

\section{Introduction}
\label{sec:intro}

Resolving lexical ambiguity remains one of the most challenging
problems in natural language processing. It is often studied as a word
sense disambiguation (WSD) problem, which is the task of assigning the correct sense to a word in a given context \cite{senseval:kilgariff:1998}.
Word sense disambiguation is typically tackled using only \emph{textual context}; however, in a multimodal setting, \emph{visual context} is also
available and can be used for disambiguation. Most prior work on
visual word sense disambiguation has targeted noun senses
\cite{Barnard:isd:2005:backup,loeff:isd:acl:2006,saenko:isd:nips:2009},
but the task has recently been extended to verb senses \cite{GellaLK16,GellaKL19}.
Resolving sense ambiguity is particularly crucial for translation
tasks, as words can have more than one translation, and these
translations often correspond to word senses
\cite{carpuat2007wsdmt,Roberto:2009}.  As an example consider the verb
\textit{ride}, which can translate into German as \textit{fahren}
(ride a bike) or \textit{reiten} (ride a horse). Recent work on
multimodal machine translation has partly addressed lexical ambiguity
by using visual information, but it still remains unresolved especially for the part-of-speech categories such as verbs \citep{specia2016shared,Shah:shef:mmt:2016,imagePivotACL16,lala2018mlt}. Prior work on
cross-lingual WSD has been limited in scale and has only employed
textual context \cite{crosslingual:wsd:semeval:2013}, even though the
task should benefit from visual context, just like monolingual WSD.

\begin{figure}[t]
\begin{center}\includegraphics[height=30mm]{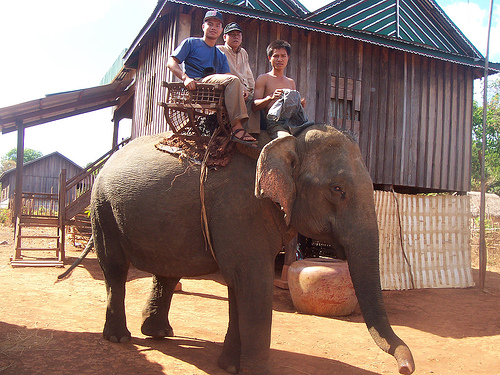}\end{center}
\setlength{\tabcolsep}{2pt}
\begin{tabular}{ll}
Source: & Three guys \textbf{riding} on an elephant. \\
Target: & Drei M\"{a}nner \textbf{reiten} auf einem Elefanten. \\
Output: & Drei Jungs \textcolor{red}{\textbf{fahren}} auf einem Elefanten.
\end{tabular}
\caption{An example of a verb sense translation error (shown in \textcolor{red}{\textbf{bold red}}) by the English-German neural translation system of \citeauthor{sennrichwmt2017} (\citeyear{sennrichwmt2017}).} 
\label{fig:wmt17-en-de-translations}
\end{figure}

Visual information has been shown to be useful to map words across languages for bilingual lexicon induction. For this, images are used
as a pivot between languages or visual information is combined with cross-lingual vector spaces to learn word translations across languages
\cite{bergsma2011learning,kiela2015visual,vulic2016multi}.  However,
as with other grounding or word similarity tasks, bilingual lexicon
induction has so far mainly targeted nouns and these approaches was shown to perform poorly for other word categories such as verbs. Recent work by \citet{GellaSKL17} and \citet{KadarECCA18} has shown using image as pivot between languages can lead to better multilingual multimodal representations and can have successful applications in crosslingual retrieval and multilingual image retrieval.

In this paper, we introduce the \ourdataset dataset of 9,504 images
annotated with English verbs and their translations in German and
Spanish. For each image in \ourdataset, the English verb is
translation-ambiguous, i.e., it has more than one possible translation
in German or Spanish. We propose a series of disambiguation models
that, given an image and an English verb, select the correct
translation of the verb. We apply our models on \ourdataset and find
that multimodal models that fuse textual context with visual features
outperform unimodal models, confirming our hypothesis that
cross-lingual WSD benefits from visual context.

Cross-lingual WSD also
has a clear application in machine translation. Determining the correct sense of a verb is important for high quality translation output, and sometimes text-only translation
systems fail when the correct translation would be obvious from visual
information (see Figure~\ref{fig:wmt17-en-de-translations}). To show that cross-lingual visual sense disambiguation can improve the
performance of translation systems, we annotate a part of our
\ourdataset dataset with English image descriptions and their German
translations. 
There are two existing multimodal translation evaluation sets with ambiguous words: the Ambiguous COCO dataset \cite{ElliottFBBS17} contains sentences that are ``possibly ambiguous'', and the Multimodal Lexical Translation dataset is restricted to predicting single words instead of full sentences \cite{lala2018mlt}. This type of resource is important for multimodal translation because it is known that humans use visual context to resolve ambiguities for nouns and gender-neutral words \cite{frank_elliott_specia_2018}. \ourdataset contains sentences that are known to have ambiguities, and it allows for sentence-level and verb prediction evaluation. Here, we use the verbs predicted by our visual sense
disambiguation model to constrain the output of a neural translation system and
demonstrate a clear improvement in Meteor, BLEU, and verb accuracy
over a text-only baseline.


\section{\ourdataset Dataset} \label{sec:dataset}

\paragraph{Images Paired with Verb Translations}
The \ourdataset dataset pairs sense-ambiguous English verbs with images as visual context and contextually appropriate German and Spanish translations. Table \ref{tab:dataset-examples} shows examples of images taken from \ourdataset with their Spanish and German translations. To compile the dataset, we first chose a set of English verbs which had multiple translations into German and Spanish in
Wiktionary, an online dictionary. Then we retrieved 150 candidate images from Google Images using queries that included the target English verb. We constructed the verb phrases by extracting the 100 most frequent phrases for each verb from the English Google syntactic n-grams dataset \cite{lin2012syntactic}, which we then manually filtered to remove redundancies, resulting in 10 phrases per verb. Examples of verb phrases for \textit{blow} include \textit{blowing hair}, \textit{blowing a balloon}, and \textit{blowing up a bomb}. We filtered the candidate images using crowdworkers on Amazon Mechanical Turk, who were asked to remove images that were irrelevant to the verb phrase query. Overall pairwise agreement for this image filtering task was 0.763. Finally, we employed native German and Spanish speakers to translate the verbs into their language, given the additional visual context.

This resulted in a dataset of 9,504 images, covering 55 English verbs with 154 and 136 unique translations in German and Spanish, respectively. We divided the dataset into 75\% training, 10\% validation and 15\% test splits.

\paragraph{Sentence-level Translations}
We also annotated a subset of \ourdataset with sentence-level translations for English and German. This subset contains 995 image--English description--German translation tuples that can be used to evaluate the verb sense disambiguation capabilities of multimodal translation models. We collected the
data in four-steps: (1)~crowdsource English descriptions of the images using the gold-standard \ourdataset verb as a prompt; (2)~manually post-edit the English descriptions to ensure they contain
the correct verb; (3)~crowdsource German translations, given the English descriptions, the German gold-standard \ourdataset verb, and the image; (4)~manually post-edit the German translations to ensure they contain the correct verb. Figure \ref{fig:wmt17-en-de-translations} shows an example of an image paired with its English description and German translation.

\begin{table}
\centerline{%
\begin{tabular}{l@{}c@{}c@{}c}
& \includegraphics[height=20mm, width = 20mm]{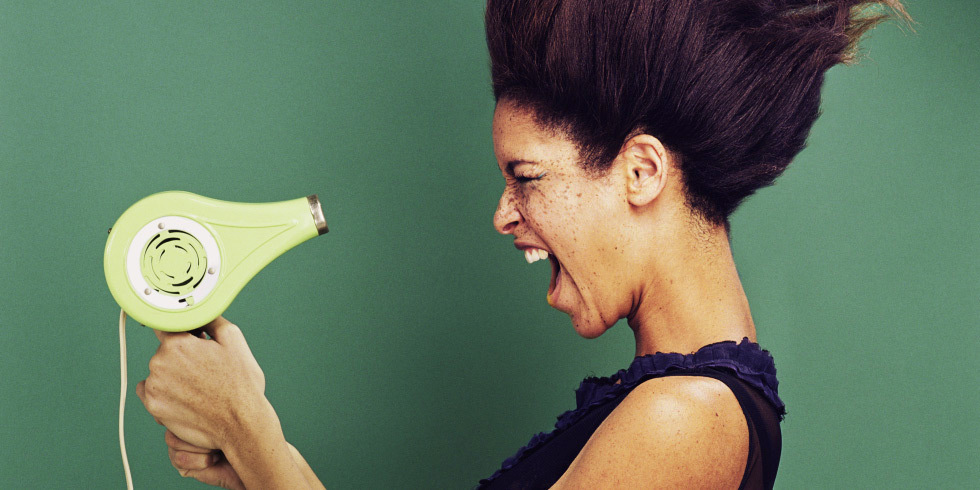} &
\includegraphics[height=20mm, width = 20mm]{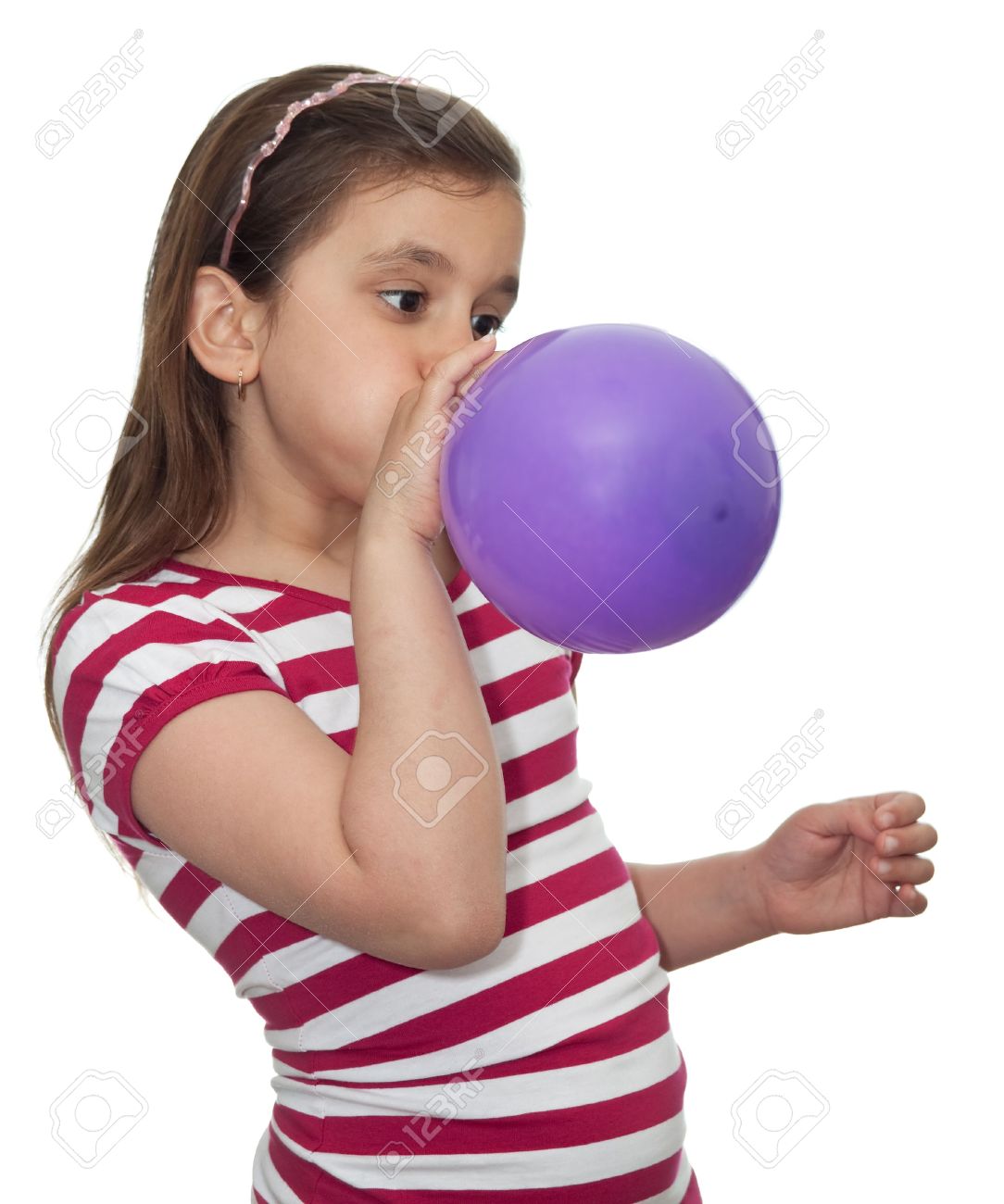} & 
\includegraphics[ height=20mm, width = 20mm]{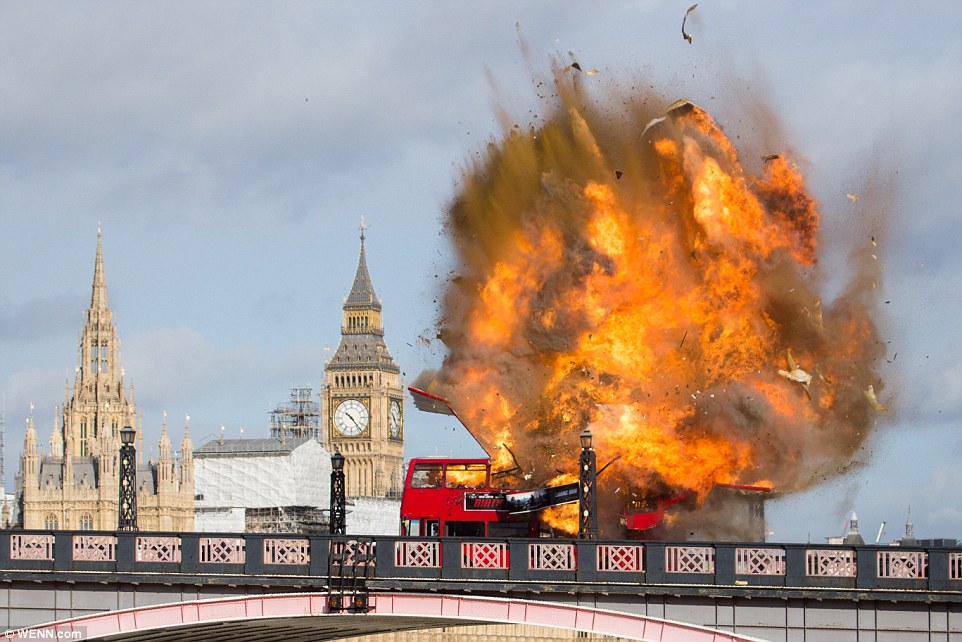} \\[1ex]
\toprule
Spanish & mandar  & hinchar & explotar\\
German  & zublasen & aufblasen & detonieren\\
\bottomrule
\end{tabular}
}
\caption{Images for the English verb \textit{blow}
  annotated with translations in Spanish and German. The images
  correspond to the uses of \textit{blowing
    with a hair dryer} and \textit{blowing a balloon}, and \textit{blowing up a bomb}.}
\label{tab:dataset-examples}
\end{table}




\section{Verb Sense Disambiguation Modeling}
\label{sec:modeling}

We propose three models for cross-lingual verb sense disambiguation, based on the visual input, the textual input, or using both inputs. Each model is trained to minimize the negative log probability of predicting the correct verb translation.

\subsection{Unimodal Visual Model}
\label{sec:visual_class}

Visual features have been shown to be useful for learning semantic representations of
words \cite{angeliki:multi-skipgram:2015}, bilingual lexicon learning
\cite{kiela2015visual}, and visual sense disambiguation \cite{GellaLK16}, amongst others. We propose a model that learns to predict the verb translation using only visual input. Given an image \textbf{I}, we extract a fixed feature vector from a Convolutional Neural Network, and project it into a hidden layer \textbf{h$_v$} with the learned matrix \textbf{W$_i$} $\in \mathbb{R}^{h \times 512}$ (Eqn. \ref{eqn:vhid}). The hidden layer is projected into the output vocabulary of $v$ verbs using the learned matrix \textbf{W$_o$} $\in \mathbb{R}^{h \times v}$, and normalized into a probability distribution using a softmax transformation (Eqn. \ref{eqn:vout}).
\begin{align}
 \textbf{h$_v$} &= \textbf{W$_i$} \cdot \texttt{CNN}(\textbf{I}) + \textbf{b$_i$} \label{eqn:vhid} \\
 \textbf{y} &= \texttt{softmax}(\textbf{W}_o \cdot \textbf{h$_v$} + \textbf{b$_o$}) \label{eqn:vout}
\end{align}

\subsection{Unimodal Textual Model}
\label{sec:textual-classifiers}

Each image in \ourdataset is associated with the query phrase that was used to retrieve it. Given a query phrase with $N$ words, we embed each word as a $d$-dimensional dense vector, and represent the phrase as the average of its embeddings E. We then project the query representation into a hidden layer with the learned matrix \textbf{W$_q$} $\in \mathbb{R}^{h \times d}$ (Eqn. \ref{eqn:thid}). The hidden layer is projected into an output layer and normalized to a probability distribution, in the same manner as the unimodal visual model.
\begin{align}
 \textbf{h$_q$} &= \textbf{W$_q$} \cdot \bigg(\frac{1}{N} \sum_i^N \text{E}[w_i]\bigg) + \textbf{b$_q$} \label{eqn:thid}
\end{align}

\subsection{Multimodal Model}
\label{lab=mm-classifiers}

We also propose a multimodal model that integrates the visual and textual features to predict the correct verb sense. In our multimodal model, we concatenate the inputs together before projecting them into a hidden layer with a learned matrix \textbf{W$_h$} $\in \mathbb{R}^{h \times (512+h)}$ (Eqn. \ref{eq:earlyh}).  We follow the same steps as the unimodal models to project the multimodal hidden layers into the output label space.
\begin{align}
  \textbf{h$_{early}$} &= \textbf{W$_h$} \cdot [\texttt{CNN}(\textbf{I}) \; ; \; \textbf{h$_q$}] + \textbf{b$_h$} \label{eq:earlyh} 
\end{align}


\section{Verb Disambiguation Experiments}
\label{sec:exp_and_results}

\begin{table}[t]
\renewcommand{\arraystretch}{1.0}
\tabcolsep 4pt
\centerline{%
\begin{tabular}{lcccccccc}
\toprule
 & Chance & Majority & Text & Image & MM \\ 
\midrule
German & 0.7 & 2.8 & 49.1 & 52.1 & \textbf{55.6}\\
Spanish & 0.7 & 4.0 & 52.7 & 50.3 & \textbf{56.0}\\
\bottomrule
\end{tabular}}
\caption{Cross-lingual verb sense disambiguation accuracy of our unimodal models and the multimodal model. We also show the performance of a random chance baseline and a majority label baseline.}
\label{tab:test-results}
\end{table}

Our experiments are designed to determine whether the integration of textual and visual features yields better cross-lingual verb sense disambiguation than unimodal models.

\subsection{Setup and Evaluation}
\label{sec:setup}

We embed the textual queries using
pre-trained $d=300$ dimension word2vec embeddings
\cite{word2vec:2013}. We represent images in the visual model using
the features extracted from the 512D \texttt{pool5} layer of a
pre-trained ResNet-34 CNN \cite{he2016deep}. All our models have a
$h=128$ dimension hidden layer. The German models have an output
vocabulary of $v=154$ verbs, and the Spanish models have a vocabulary
of $v=136$ verbs.  All of our models are trained using SGD with
mini-batches of 16 samples and a learning rate of 0.0001.

We evaluate the performance of our models by measuring the accuracy of the predicted verb against the gold standard. We also compare against chance and majority label baselines. Our preliminary experiments show that with better visual representation we achieve better acccuracy scores similar to others who observed better visual representation contributes to better downstream tasks such as image description \cite{msr:caption:generation:2015}, multimodal machine translation \cite{specia2016shared} and representation learning \cite{KadarECCA18}.

\begin{figure}[t]
\begin{subfigure}{0.15\textwidth}
\centering
\includegraphics[width=25mm]{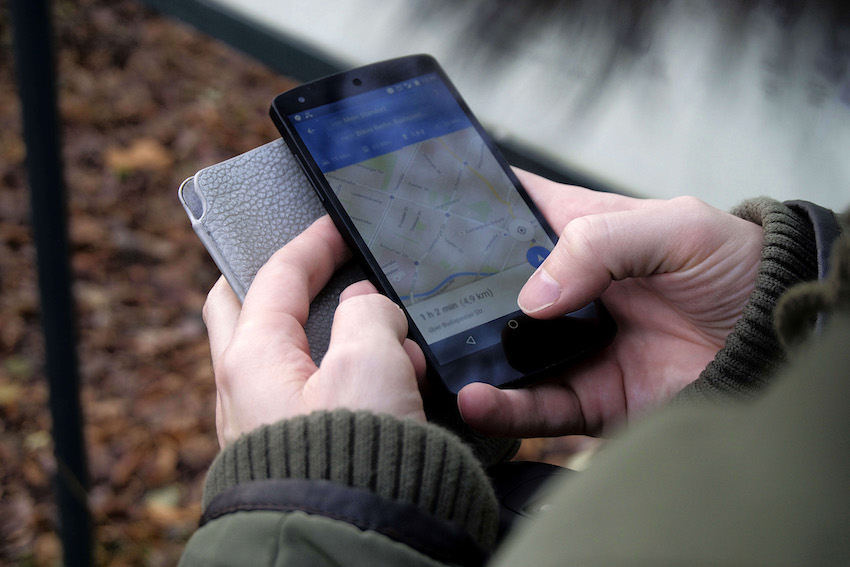} \\
{\small\textbf{looking}\\[-1ex] for directions}
\end{subfigure}
\hspace{0.5em}
\begin{subfigure}{0.2\textwidth}
\begin{tabular}{lll}
\toprule
Model & German & Spanish \\
\midrule
Textual & schauen & mirar\\
Visual & tragen & \textbf{buscar}\\
MM & \textbf{suchen} & \textbf{buscar} \\
\bottomrule
\end{tabular}
\end{subfigure}
\caption{Examples of the Top-1 predictions of our unimodal and multimodal models. Only the early fusion multimodal model predicts the correct verb sense for both languages (shown in bold).}
\label{tab:error-analysis}
\end{figure}

\begin{table*}[t]
\centering
\tabcolsep 2pt
\begin{tabular}{cccc}
\hspace{5em} & \includegraphics[height=30mm]{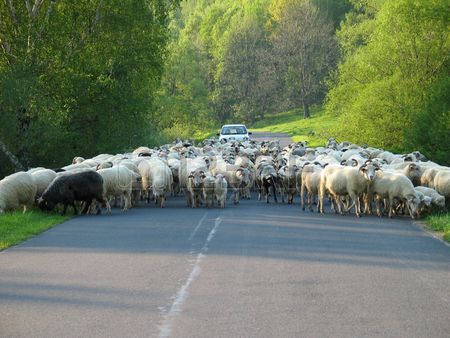} & \hspace{7em} &\includegraphics[height=30mm]{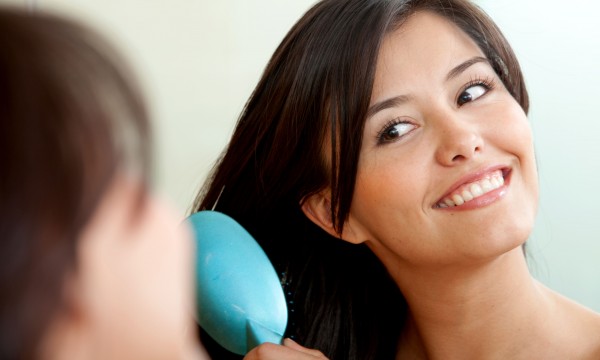} \\
\end{tabular}
\begin{tabular}{p{1.5cm}@{~}p{6cm}p{1cm}p{6cm}}
\toprule
\bf{Source} & {A large herd of sheep is \textbf{blocking} the road.}  & & {A woman smiles as she \textbf{brushes} her long, dark hair.} \\
\bf{Target} & {Eine gro\ss e Herde Schafe \textbf{blockiert} die Stra\ss e.} & &{Eine Frau l\"{a}chelt w\"{a}hrend sie sich ihre dunklen langen Haare \textbf{b\"{u}rstet} .}\\
\bf{Baseline} & {Eine gro\ss e Herde Schafe \textcolor{red}{\textbf{kriecht}} die Stra\ss e entlang.} & & {Eine Frau l\"{a}chelt , als sie ihren langen und dunklen Haaren \textcolor{red}{\textbf{putzt}} .}\\
\bf{+WSD} & {Eine gro\ss e Herde Schafe \textbf{blockieren} die Stra\ss e.} & & {Eine Frau l\"{a}chelt , als sie ihr lange , dunklen Haaren \textbf{b\"{u}rsten} .} \\
\bottomrule
\end{tabular}

\caption{The visual verb sense predictions (``blockieren'', ``b\"{u}rsten'') successfully constrains the decoder to predict the correct sense of the verb (``block'', ``brush'') in the German translation \textbf{(+WSD)}. The incorrect verb in the baseline translation is shown in \textcolor{red}{\textbf{bold red.}}}
\end{table*}

\subsection{Results}
\label{sec:results}

We present the results in Table~\ref{tab:test-results}. The chance and majority label baselines perform very poorly. The unimodal textual model performs better than the unimodal visual model for German verb sense disambiguation, but we find the opposite for Spanish unimodal verb sense disambiguation. However, the early fusion multimodal model outperforms the best unimodal model for both German and Spanish. This confirms that cross-lingual verb sense disambiguation benefits from multimodal supervision compared to unimodal supervision.

\subsection{Discussion}
\label{discussion}

We analyzed the outputs of our models in order to understand where
multimodal features helped in identifying the correct verb translation
and the cases where they failed.  In Figure \ref{tab:error-analysis}, we
show an example that illustrates how varying the input (textual,
visual, or multimodal) affects the accuracy of the verb prediction.
We show the top verb predicted by our models for both German and Spanish. The top predicted verb using text-only visual features is incorrect. The unimodal visual features model predicts the correct Spanish verb but the incorrect German verb. However, when visual information is added to textual features, models in both the languages predict the correct label.



\section{Machine Translation Experiments}
\label{sec:exp_and_results_translation}

We also evaluate our verb sense disambiguation model in the challenging downstream task of multimodal machine translation \cite{specia2016shared}. We
conduct this evaluation on the sentence-level translation subset of
\ourdataset. 
We evaluate model performance using BLEU \cite{Papineni:2002:BMA:1073083.1073135} and Meteor scores
\cite{denkowski-lavie:2014:W14-33} between the \ourdataset reference
description and the translation model output. We also evaluate the
verb prediction accuracy of the output against the gold standard
verb annotation.

\begin{table}[t]
  \centering
  \begin{tabular}{lccc}
  \toprule
               & Meteor & BLEU & VAcc \\
  \midrule
  Baseline NMT & 38.6 & 17.8 & 22.9 \\
  + Predicted Verb & 40.0 & 18.5 & 49.5 \\
  + Oracle Verb    & 40.4 & 19.1 & 77.7 \\
  \midrule
  Caglayan et al. & 46.1 & 25.8 & 29.3\\
  Helcl \& Libovick\'{y} & 42.5 & 22.3 & 25.1 \\
  \bottomrule
  \end{tabular}
  \caption{Translation results: Meteor and BLEU are standard text-similarity
    metrics; verb accuracy (VAcc) counts how often the model
    proposal contains the gold standard German
    verb.}\label{tab:translation:results}
\end{table}

\subsection{Models}
\label{ssec:translation-models}

Our baseline is an attention-based neural machine translation model \cite{Sockeye:17} trained on the 29,000 English-German sentences in \multiflickr \cite{multi30k:2016}. We preprocessed the text with punctuation normalization, tokenization, and lowercasing. We then learned a joint byte-pair-encoded vocabulary with 10,000 merge operations to reduce sparsity \cite{sennrich-haddow-birch:2016:P16-12}.

Our approach uses the German verb predicted by the unimodal visual
model (Section \ref{sec:visual_class}) to constrain the output of the
translation decoder \cite{post2018fast}. This means that our approach
does not directly use visual features, instead it uses the output of
the visual verb sense disambiguation model to guide the translation process. 

We compare our approach against two state-of-the-art multimodal
translation systems: \citet{caglayan-EtAl:2017:WMT} modulate the
target language word embeddings by an element-wise multiplication with
a learned transformation of the visual data;
\citet{helcl-libovicky:2017:WMT} use a double attention model that
learns to selectively attend to a combination of the source language
and the visual data.

\subsection{Results}

Table~\ref{tab:translation:results} shows the results of the
translation experiment. Overall, the Meteor scores are much lower than
on the \multiflickr test sets, where the state-of-the-art single model
scores 51.6 Meteor points compared to 46.1 Meteor we obtained. This
gap is most likely due evaluating the models on an out-of-domain dataset with out-of-vocabulary tokens. Using the predicted verb as a decoding constraint outperforms the text-only translation baseline by 1.4 Meteor points. In addition, the translation output of our model
contains the correct German verb 27\% more often than the text-only
baseline model. These results show that a multimodal verb sense
disambiguation model can improve translation quality in a multimodal
setting.

We also calculated the upper bound of our approach by using the gold
standard German verb as the lexical constraint. In this oracle
experiment we observed a further 0.4 Meteor point improvement over our
best model, and a further 27\% improvement in verb accuracy. This
shows that: (1)~there are further improvements to be gained from
improving the verb disambiguation model, and (2)~the OOV rate in
German means that we cannot achieve perfect verb accuracy.


\section{Conclusions}
\label{sec:concl}

We introduced the \ourdataset dataset of 9,504 images annotated with an English verb and its translation in Spanish and German. We proposed 
a range of cross-lingual visual sense disambiguation models 
and showed that multimodal models that fuse textual and visual features outperform unimodal models.  We also collected a set of image descriptions and their translations, and showed that the
output of our cross-lingual WSD system boosts the performance of a
text-only translation system on this data. MultiSense is publicly available at \url{https://github.com/spandanagella/multisense}


\section*{Acknowledgements}

DE was supported by an Amazon Research Award. This work was supported by the donation of a Titan Xp GPU by the NVIDIA Corporation.

\bibliography{references}
\bibliographystyle{acl_natbib}
\end{document}